\documentclass[runningheads]{llncs}
\usepackage[T1]{fontenc}
\usepackage{graphicx}
\usepackage[most, breakable]{tcolorbox}

\usepackage{hyperref}
\usepackage{color}
\usepackage{amsmath} 
\usepackage{booktabs} 
\usepackage{multirow} 
\usepackage{orcidlink}

\usepackage{tikz,pgfplots,pgfplotstable,xcolor}
\pgfplotsset{compat=1.18}
\usetikzlibrary{patterns}

\begin{document} 
%
\title{ReFactX: Scalable Reasoning with Reliable Facts via Constrained Generation\thanks{This preprint has not undergone peer review or any post-submission improvements or corrections. The Version of Record of this contribution was published in \textit{The Semantic Web -- ISWC 2025} and is available online at \url{https://doi.org/10.1007/978-3-032-09527-5_16}.}}
%
\titlerunning{ReFactX: Scalable Reasoning with Facts via Constrained Generation}
%
\author{Riccardo Pozzi\inst{1,3}\orcidlink{0000-0002-4954-3837}\thanks{This work was done while Riccardo Pozzi was visiting InfAI and TU Dresden.} \and
Matteo Palmonari\inst{1}\orcidlink{0000-0002-1801-5118} \and
Andrea Coletta\inst{2}\orcidlink{0000-0003-1401-1715}\thanks{The opinions expressed are personal and should not be attributed to Banca d'Italia.} \and
Luigi Bellomarini\inst{2}\orcidlink{0000-0001-6863-0162}$^{\star\star}$ \and
Jens Lehmann\inst{3,5,7}\orcidlink{0000-0001-9108-4278}\thanks{This work was done outside of Amazon.} \and
Sahar Vahdati\inst{3,4,5,6}\orcidlink{0000-0002-7171-169X}
}

\authorrunning{R. Pozzi et al.}

\institute{University of Milano-Bicocca, 20126 Milano, Italy \\
\email{riccardo.pozzi@unimib.it} \and
Banca d'Italia, 00184 Roma, Italy \and
Institute for Applied Informatics (InfAI), 04109 Leipzig, Germany \and
TIB Leibniz Information Centre for Science and Technology, 30167 Hannover, Germany \and
ScaDS.AI Dresden/Leipzig, Technische Universität Dresden, 01069 Dresden, Germany \and
Data Science Institute, Leibniz University Hannover, 30167 Hannover, Germany \and
Amazon, 01097 Dresden, Germany
}
\institute{University of Milano-Bicocca, Milano, Italy \\
\email{riccardo.pozzi@unimib.it} \and
Banca d'Italia, Roma, Italy \and
Institute for Applied Informatics (InfAI), Leipzig, Germany \and
TIB Leibniz Information Centre for Science and Technology, Hannover, Germany \and
ScaDS.AI Dresden/Leipzig, Technische Universität Dresden, Dresden, Germany \and
Data Science Institute, Leibniz University Hannover, Hannover, Germany \and
Amazon, Dresden, Germany
}

\maketitle              
\begin{abstract}
Knowledge gaps and hallucinations are persistent challenges for Large Language Models (LLMs), which generate unreliable responses when lacking the necessary information to fulfill user instructions.
Existing approaches, such as Retrieval-Augmented Generation (RAG) and tool use, aim to address these issues by incorporating external knowledge. Yet, they rely on additional models or services, resulting in complex pipelines, potential error propagation, and often requiring the model to process a large number of tokens.
In this paper, we present a scalable method that enables LLMs to access external knowledge without depending on retrievers or auxiliary models. Our approach uses constrained generation with a pre-built prefix-tree index. Triples from a Knowledge Graph are verbalized in textual facts, tokenized, and indexed in a prefix tree for efficient access. During inference, to acquire external knowledge, the LLM generates facts with constrained generation which allows only sequences of tokens that form an existing fact.
We evaluate our proposal on Question Answering and show that it scales to large knowledge bases (800 million facts), adapts to domain-specific data, and achieves effective results. These gains come with minimal generation-time overhead. ReFactX code is available at https://github.com/rpo19/ReFactX.


\end{abstract}




%
%
%
\section{Introduction}



Large Language Models (LLMs) have demonstrated a variety of capabilities, ranging from natural language understanding and generation to reasoning, especially when enhanced with techniques like Chain-of-Thought (CoT)~\cite{wei2022chainofthought} prompting. However, LLMs are prone to hallucinations~\cite{maynez-etal-2020-faithfulness-hallucination} and their internal knowledge remains limited to their training data~\cite{cheng2024dateddatatracingknowledge,li2024knowledgeboundarylargelanguage}. This complicates their application to knowledge-intensive tasks like Question Answering (QA), especially when the tasks require accessing information not available in LLM parametric knowledge, such as time-critical data or corporate data.

Retrieval-Augmented Generation (RAG)~\cite{lewis2020retrievalaugmented,xu2023retrieval} and tool-use~\cite{yao2023reactsynergizingreasoningacting,thoppilan2022lamdalanguagemodelsdialog,lehmann2024beyond} represent two prominent families of solutions. They enable LLMs to consult external Knowledge Bases (KBs) or execute external tools to enrich their responses with grounded information. We will refer to these methods as Knowledge-Enhanced QA (KE-QA). They often rely on additional components such as retrievers or Entity Linking systems, or auxiliary models, which are orchestrated in information processing pipelines.
However, such pipeline-based or service-based solutions have a number of drawbacks. They require the optimization of a larger number of parameters and make it difficult to effectively back-propagate supervision signals to early components of the pipeline~\cite{zhou2025openragoptimizingragendtoend}. These approaches are susceptible to error-propagation~\cite{le-fokkens-2017-tackling}, and knowledge updates may be complex to propagate to all the different components (e.g., to QA and Entity Linking knowledge bases).

An interesting technique that may overcome such limitations is constrained generation~\cite{cao2021autoregressive,scholak-etal-2021-picard}, which restricts the model's output space during decoding to sequences that satisfy predefined structural, syntactic, or semantic constraints. A few recent approaches applied constrained generation on KE-QA, achieving promising results~\cite{li2024decodinggraphsfaithfulsound,luo2024graphconstrainedreasoningfaithfulreasoning}.
However, as the application of this technique to KE-QA is quite novel, there is an important research challenge to address to advance these techniques: the scalability to large Knowledge Bases. Can these approaches be applied to large KBs such as Wikidata\footnote{https://www.wikidata.org}?
In this paper, we present Reliable Fact eXtractor (ReFactX), an approach that addresses this challenge and poses itself as an alternative to pipeline-based solution.
%
%
ReFactX allows LLMs to integrate external knowledge without depending on additional models or external retrievers, but relying solely on constrained generation, supported by a pre-built prefix-tree index, designed to facilitate and speed-up fact retrieval.

At inference time, the LLM is instructed via In-Context Learning (ICL)~\cite{dong2024survey} to invoke the \emph{Fact} command when external facts are required. Once the command is recognized, constrained generation is activated: the model produces tokens only along valid paths in the prefix tree, guaranteeing that the output matches a fact from the Knowledge Base. Once an entire KB fact is generated, the decoding mechanism reverts back to normal.
It is important to note that every generated token is selected based on LLM's probability estimates. Constrained generation only narrows the vocabulary to the tokens that form an existing fact in the KB, but always leaves the final choice to the LLM.

\begin{figure}[h]
    \centering
    \includegraphics[width=\textwidth]{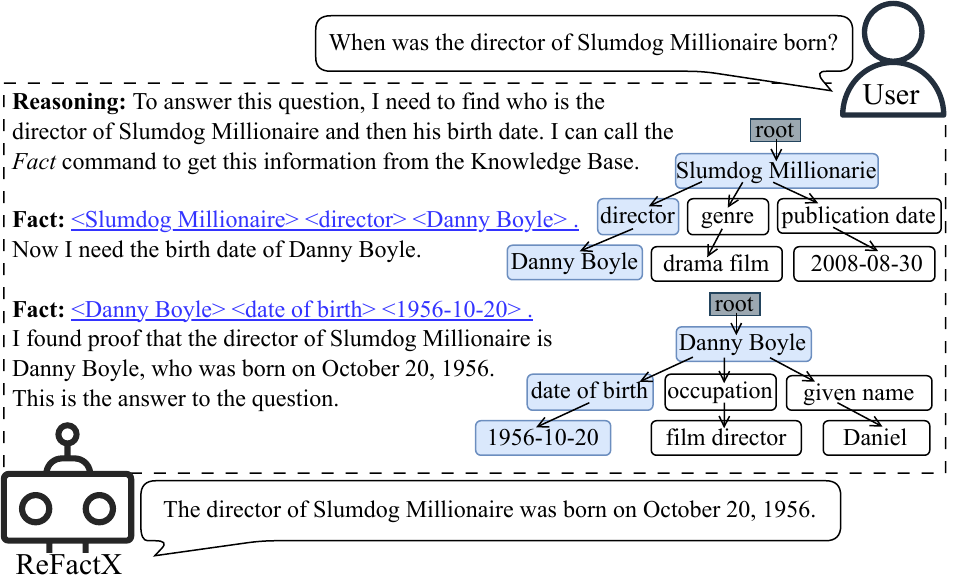}
    \caption{\textbf{ReFactX Answering an Open-domain Question.}
The LLM sketches a plan, makes two \emph{Fact} calls, and---with constrained generation (blue underline)---inserts valid facts from a Wikidata-based prefix tree before giving the final answer.
    }
    \label{fig:example}
    \end{figure}

\textbf{Motivating Example.} We illustrate our approach in Figure~\ref{fig:example}, by depicting an example with the question \textit{``When was the director of Slumdog Millionaire born?''}. The model is instructed to first reason on how to reach the correct answer, and then it starts to acquire facts.
Once the \emph{Fact} command is called, we enable constrained generation (underlined in blue) and we guide the model to generate an existing fact. This way the model is able to first find that ``Danny Boyle'' is the director of ``Slumdog Millionaire'', then to find his birth date, and finally to answer the question correctly.


We can summarize the main contributions of this work as follows:
\begin{itemize}
  \item \textbf{ReFactX}: a generic, constrained-generation wrapper that lets any LLM access very large KBs, with no external retrievers or pipelines.  
  \item \textbf{Efficient and scalable}: a disk-backed prefix tree holds up to 800M facts and adds only $\sim$1\,\% latency.  
  \item \textbf{Empirical validation}: insights on four QA benchmarks suggest that ReFactX can achieve competitive results and obtains accuracy gains of up to 20\% at over 90\% precision when compared to LLMs answering only with their parametric knowledge.
\end{itemize}

\section{Related Work}
\label{sec:rw}

Several approaches have been proposed to incorporate external knowledge into LLMs~\cite{gao2024retrievalaugmentedgenerationlargelanguage}. In our work, we focus on Question Answering (QA)---more specifically, on \textbf{Knowledge-Enhanced QA (KE-QA)}, where a model acquires external knowledge for answering questions.

Most of approaches can be classified as \textbf{input-based KE-QA}.
They rely on external tools, such as search engines or dense retrievers~\cite{karpukhin-etal-2020-dense}, which can be arranged in pipeline-based workflows~\cite{lewis2020retrievalaugmented,xu2023retrieval} or actively called by LLMs~\cite{yao2023reactsynergizingreasoningacting,NEURIPS2023_d842425e,lehmann2024beyond}. These approaches usually feed external knowledge as input (either in plain text or with dense embeddings) to the LLMs. Input-based KE-QA approaches, while effective in reducing hallucinations~\cite{shuster-etal-2021-retrieval-augmentation}, require additional models or services for retrieving external information that can introduce delays, especially when calling remote APIs, increase training complexity~\cite{sun2025zerosearchincentivizesearchcapability},
and raise substantially the number of input tokens~\cite{yang2024memory3}, leading to higher response time and resource usage.

Other approaches can be classified as \textbf{memory-based KE-QA}. They propose memory modules, separated from model parameters, which make the models' internal mechanisms interact with external knowledge, e.g., by using cross-attention to incorporate external vectors~\cite{yang2024memory3,wu-etal-2022-efficient}. While memory-based KE-QA approaches fuse external knowledge into LLMs' generation process more deeply, they typically require architectural modifications, complicating the use of pretrained LLMs.


Constrained generation~\cite{cao2021autoregressive,scholak-etal-2021-picard}, described in detail in Section~\ref{sec:constrained-gen}, is a promising technique to avoid the drawbacks of input-based and memory-based KE-QA approaches. It has been successfully applied to various tasks, including code generation~\cite{scholak-etal-2021-picard}, so that the model always produces syntactically correct code, Entity Linking~\cite{cao2021autoregressive} or Information Retrieval~\cite{bevilacqua2022autoregressive}. 
It has also been applied to instruction-following LLMs for several tasks that require the LLM to produce a specifically structured output, such as Entity Disambiguation~\cite{geng-etal-2023-grammar}.

Indeed, two recent approaches~\cite{li2024decodinggraphsfaithfulsound,luo2024graphconstrainedreasoningfaithfulreasoning} proposed \textbf{constrained generation for KE-QA}, especially for guiding LLMs through paths from a Knowledge Graph (KG). However, they do not address the problem of accessing facts from large knowledge bases in a scalable way (800M facts, in our case),  and still rely on entity linking systems for extracting question entities, similarly to input-based approaches. 
A first approach, \textit{Decoding on Graphs (DoG)}~\cite{li2024decodinggraphsfaithfulsound}, generates triples from a Knowledge Graph in a reasoning process similar to CoT but grounded on KG knowledge, also allowing the model to alternate the constrained generation of triples and the reasoning with normal generation.
A second approach, \textit{Graph-Constrained Reasoning (GCR)}~\cite{luo2024graphconstrainedreasoningfaithfulreasoning} generates entire paths from a KG using a fine-tuned LLM to find answer hypotheses, and then asks a top-tier language model to give a final answer based on the paths.

\section{ReFactX: Enabling LLMs to Access Large-Scale KBs}
\label{sec:approach}
ReFactX allows LLMs to efficiently access large KBs at inference time with constrained generation, generating valid facts, and weave them into chain-of-thought reasoning. 
The three subsections below walk through (i) the decoding mechanism, (ii) the scalable Wikidata index, and (iii) how both pieces plug into a QA workflow.




\subsection{Constrained Fact-Generation Mechanism}
\label{sec:constrained-gen}

The mechanism of constrained generation alters the autoregressive next-token generation process, in which the LLM parameterized by $\theta$, iteratively estimates the probability of each token $P(t)$ from the vocabulary $V$ to be the chosen as the next-token, given the current sequence $t_0..t_k$:
\begin{equation}
    P_\theta(t | t_{0}..t_{k}) \forall t \in V
\end{equation}
The next-token $t_{k+1}$ can be chosen according to different sampling strategy;
in greedy decoding the most probable token is chosen as follows:
\begin{equation}
    t_{k+1} = \text{argmax}_{t \in V}P_\theta(t | t_{0}..t_{k})
    \label{eq:token_selection}
\end{equation}

\begin{figure}[h]
    \centering
    \fbox{%
    \parbox{.93\textwidth}{%
    \begin{tabular}{ll}
        \multicolumn{2}{l}{\textbf{When was Danny Boyle born?}} \\
        \textcolor{red}{Normal:} & \textcolor{red}{<Danny Boyle> <born> <1960-09-15> .} \\
        \textcolor{blue}{Constrained:} & \textcolor{blue}{\underline{<Danny Boyle> <date of birth> <1956-10-20> .}}
    \end{tabular}
    \hspace*{2em}\includegraphics{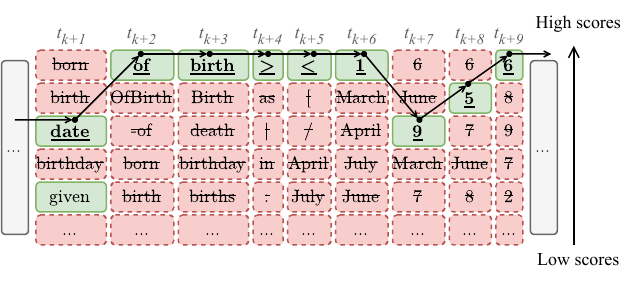}
    }
    }
    \caption{\textbf{Constrained decoding steers the LLM toward the correct fact.}
At each step constrained decoder chooses the highest-probability token that still completes a Wikidata fact, avoiding invalid branches and yielding ``<Danny Boyle> <date of birth> <1956-10-20> .''.}
    \label{fig:constrained_generation}
\end{figure}

Intuitively, to constrain this process for generating only existing facts, we need to restrict $V$ to only allow tokens that can form a fact from the KB. We define $V^A_{t_0..t_k}$, which contains all the tokens that can lead to an existing fact if added to the sequence $t_0..t_k$. Considering the example in Figure~\ref{fig:constrained_generation}, at step $t_{k+1}$, when $t_0..t_k=\text{``<Danny Boyle> <''}$, $V^A_{t_0..t_k} = \{\text{date}, \text{given}\}$.

Next, we define the \textit{next tokens} function $NT_{KB}$ that, given $t_0..t_k$, obtains $V^A_{t_0..t_k}$ for the considered KB.
%
Consequently, the token selection formula from Eq.~\ref{eq:token_selection} is updated to:
\begin{equation}
    t_{k+1} = \text{argmax}_{t \in V^A_{t_0..t_k}}P_\theta(t | t_{0}..t_{k})\text{, where }
    V^A_{t_0..t_k} = NT_{KB}(t_0..t_k).
\end{equation}

However, instead of modifying the vocabulary, the same result can be achieved by altering the probability distribution, setting the probability to zero for all the forbidden tokens:
\begin{equation}
    P^{c}_\theta(t | t_0..t_k) = \begin{cases}
        P_\theta(t | t_0..t_k) & \text{if } t \in V^A_{t_0..t_k}, \\
        0 & \text{otherwise}.
    \end{cases}
\end{equation}
By using $P^{c}_\theta$, instead of $P_\theta$, we achieve constrained generation, relying on the assumption that the sampling strategy (Equation~\ref{eq:token_selection}) would never choose any $t_{k+1}|P^c(t_{k+1}) = 0$.
In practice, since at implementation level the models use log-probabilities, we directly set  the log-probabilities of forbidden tokens to $-\infty$\footnote{We use Huggingface LogitsProcessor to alter the log-probabilities.}:
\begin{equation}
    \log P^{c}_\theta(t | t_0..t_k) = \begin{cases}
        \log P_\theta(t | t_0..t_k) & \text{if } t \in V^A_{t_0..t_k}, \\
        -\infty & \text{otherwise}.
    \end{cases}
\end{equation}


The application of constrained generation is depicted in Figure~\ref{fig:constrained_generation}, where we show a fact generated by the same LLM 
for the question \textit{``When was Danny Boyle born?''} first in normal generation mode (in red) and then with constrained generation (underlined in blue). While, the fact generated in normal mode is not correct, with constrained generation we are able to guide the LLM to the correct fact. The lower part of the figure details the mechanism of constrained generation. For each decoding step, allowed tokens are displayed in green boxes, whereas forbidden tokens are displayed in red boxes with dashed borders and a strikethrough.
Tokens $t_k$ are arranged in ascending order according to $P_\theta(t_k)$, from bottom to top.

Starting from the sequence 
$t_0..t_k = \text{``<Danny Boyle> <''}$, normal generation would generate $t_{k+1} = \text{argmax}_{t \in V}P_\theta(t | t_{0}..t_{k}) = \text{``born''}$, leading to an incorrect fact, while with constrained generation $t_{k+1} = \text{argmax}_{t \in V}P^c_\theta(t | t_{0}..t_{k}) = \text{``date''}$.
This happens because in the KB there is no fact starting with ``<Danny Boyle> <born''.
Subsequently, when generating $t_{k+7}$, constrained generation mechanism guides the model to select ``9'' that leads to the correct birth year of Danny Boyle ``<Danny Boyle> <born> <1956'', avoiding the model to generate ``<Danny Boyle> <born> <16'' or ``<Danny Boyle> <born> <196'', both leading to incorrect information.
This mechanism is additionally improved by beam search~\cite{hu-etal-2015-improved} which allows the model to explore in parallel multiple paths in the prefix tree.

\subsection{Scaling to 800M Facts from Wikidata}
\label{sec:wikidata-repr}


We start from the Wikidata truthy dump (Wikidata’s highest-confidence statements, excluding qualifiers\footnote{\url{https://www.wikidata.org/wiki/Wikidata:Database_download/en\#RDF\_dumps}}) of the triples\footnote{We use the dump from 11 December 2024.}.
To filter out uninformative facts we keep only the triples whose subject and relation have Wikidata identifiers, and for the object, we allow only entities with Wikidata identifiers or English literals (also numbers, dates, and literals with no language).
Then, for each entity, we obtain a meaningful textual label (as Wikidata IDs lack meaning for LLMs) corresponding to the Wikipedia title if the entity is described in English Wikipedia~\footnote{https://en.wikipedia.org/}. Otherwise we use the template \verb|rdf-schema#label (schema.org/description)|, e.g., ``Jane Hajduk (American actress)'', since the label alone is not unique.
With this process we obtain more than 800M facts like the ones underlined in Figure~\ref{fig:example}. 

Now we tokenize the facts and we index them in a prefix tree for efficient access. Additionally, while creating the tree, we calculate the number of reachable leaves from each node to avoid that LLMs generate the same fact repeatedly, a behavior we witnessed during preliminary experiments.

\begin{figure}[h]
    \centering
    \includegraphics[width=.8\linewidth]{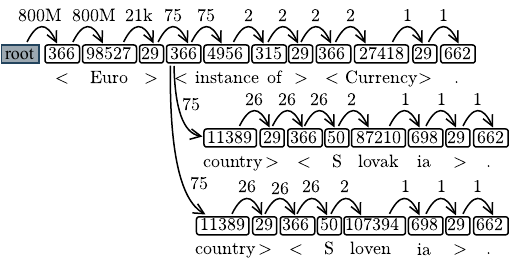}
    \caption{\textbf{Generating facts from the fact tree. } Token ids are surrounded by rounded rectangles. The arrows $S \xrightarrow{n} t_{k+1}$ represent the selection of the next token $t_k$ from the current sequence $S=t_0..t_k$ and $n$ is the number of leaves reachable from $S$.}
    \label{fig:trie_example}
\end{figure}

Duplicate facts generation is prevented as follows. Considering the prefix $X = \text{``<Euro> <country> <S''}$ in Figure~\ref{fig:trie_example}, we see that $numleaves(X)=2$. After generating $\text{``<Euro> <country> <Slovakia> .''}$, this number will be decreased to $1$ and from $X$ the model will only be allowed to select $\text{``lovenia> .''}$. At this point, all the leaves reachable from $X$ have been already generated ($numleaves(P)=0$), and generating ``S'' from $X'=\text{``<Euro> <country> <''}$ is forbidden by constrained generation. From $X'$ the LLM can generate one of the remaining $numleaves(X') - 2 = 24$ facts (e.g., $\text{``<Euro> <country> <Italy> .''}$). 

While relatively small trees can be kept in memory, e.g., using Python dictionaries, when dealing with bigger trees, such as the 800M facts tree derived from Wikidata, this is often impossible. Therefore, we rely on solid relational database software. We use PostgreSQL\footnote{https://www.postgresql.org/} (version 17.2), designing the fact-tree table in such a way that all the information required for generating a token is quickly available.
Given a prefix $X$, this includes the allowed next tokens $T = t^0..t^n$, the number of leaves reachable from $X$, and the number of reachable leaves if choosing $t$, for each $t \in T$ (to avoid selecting 0-leaves tokens and generating a fact twice). In Table~\ref{tab:postgres-table}, they are depicted respectively as \textit{Next Tokens} and \textit{\#Lv. (number of leaves)}, and \textit{Child.\#Lv.}.
However, this tree-representation schema can grow quickly in disk space required because a single path of length $L$ requires $L-1$ rows to be represented. To reduce the disk space requirements, we apply two additional measures. First, after a certain prefix length $L_c$ we stop adding rows and instead we directly save the Python subtree in Pickle format\footnote{https://docs.python.org/3/library/pickle.html}, as shown in the last row of Table~\ref{tab:postgres-table}. List items are saved as PostgreSQL arrays, while Pickle data is saved in \textit{BYTEA}.


In this way, during inference, subtrees that are manageable in size are directly loaded in memory (e.g., with $L_c = 7$, $99\%$ of the subtrees use less than 116KB).
Secondly, we represent single-leaf sequences in a single row, as visible at the fourth row in Table~\ref{tab:postgres-table}, saving the rest of the sequence in \textit{Child.\#Lv.} (the token 29 comes after 694, and 662 is the last in the sequence). Finally, for fast access, the \textit{Prefix} is indexed with a B-Tree\footnote{\url{https://www.postgresql.org/docs/current/btree.html}} index that provides logarithmic search time~\cite{cormen2022introduction}.
The ingestion process consists in creating the tree in-memory and then persisting it in the database. However, hardware memory limits do not allow to keep the entire tree in-memory. So we perform this process in batches, constructing a limited-size tree, ingesting it, and then proceeding with the next one. This method allows to process the entire 800M facts, but results in duplicated prefix rows. Indeed, same prefixes from different batches are added as different rows and need to be merged together when retrieved during inference. The maximum number of duplicates corresponds to the number of batches used during the ingestion process.
With this mechanisms we are able to index the 800M facts using 95GB.
Note that the index, storing token ids, depends on the LLM vocabulary, thus LLMs using distinct vocabularies require separate indexes.

\begin{table}[htbp]
    \centering
    \caption{PostgreSQL table content. \textit{\#Lv.} stands for ``number of leaves''.}
    \begin{tabular}{l|l|l|l|l}
    \toprule
    \textbf{Prefix} & \textbf{Next Tokens} & \textbf{\#Lv.} & \textbf{Child.\#Lv.} & \textbf{Subtree} \\
    \midrule
    \{\verb|root|\} & \{366,1134,...\} & 5M & \{5M,3,...\} & \\
    \{\verb|root|\} & \{366,8730,...\} & 5M & \{5M,9,...\} & \\
    \{\verb|root|,366\} & \{537,7350,...\} & 2M & \{2M,7,...\} & \\
    \{\verb|root|,694\} & \{29,...,662\} & 1 & \{\} & \\
    \{\verb|root|,366,...\} & \{21538,4168\} & 7 & \{4,3\} & \verb|\x804| \\
    \bottomrule
    \end{tabular}
    \label{tab:postgres-table}
    \end{table}

\subsection{Embedding ReFactX into Question-Answering Workflows}
\label{sec:refactx-qa}

\begin{figure}[ht!]
    \centering
\begin{tcolorbox}[colframe=black!70, colback=gray!10, title=] 
You are a helpful question-answering assistant that bases its answers on facts from a knowledge base and always respects the prompt.

\vspace{4pt}

The process to answer questions:

\vspace{4pt}

{\leftskip=2em
    You receive an input question.

    \vspace{4pt}

    You determine the reasoning path needed to answer the question based on the information available.

    \vspace{4pt}

    You determine the kind of answer you are asked. It can be a yes/no, a single entity, or a list of entities. Pay attention to the questions whose answer is a list of entities (e.g. Which countries share a border with Spain?): you need to find all the answer entities and include them all in the final answer.

    \vspace{4pt}

    You get relevant facts with the \textbf{``Fact:''} command. You can rely on these facts and use them a proof for your answer.
    While getting facts you continue the reasoning explaining it step by step.

    \vspace{4pt}

    Often description or short description may be useful for answering questions.

    \vspace{4pt}

    You conclude with a concise answer that depending on the question can be a yes/no, a single entity, or a list of entities. Pay attention to the questions whose answer is a list of entities.

    \vspace{4pt}

    The answer MUST be based on the proofs you found with \textbf{``Fact:''}.

    \vspace{4pt}
    
}

If you didn't find proofs with \textbf{``Fact:''} that support an answer you stop and you reply: ``I don't know.''.

\vspace{4pt}

If the question requires to find proof that an event happen and you didn't find any proof, you can assume that event didn't happen.

\vspace{4pt}

In case you are taking too long for answering (e.g., you already generated ten facts that are not useful for the question), you stop and you answer based on the proofs you acquired to that point.

\vspace{4pt}

You must always follow these instructions precisely and ensure your responses adhere strictly to this prompt.

 \end{tcolorbox}
\caption{\textbf{System prompt for the Wikidata KB.} The prompt instructs the LLM to reason step-by-step, issue \emph{Fact} calls to access facts from the Wikidata fact tree, and deliver an answer only after evidence is gathered—otherwise respond ``I don’t know.''}
\label{fig:prompt}
 
 \end{figure}

ReFactX relies on In-Context Learning (ICL)~\cite{dong2024survey} for instructing the models to access external knowledge.
Our prompt (in Figure~\ref{fig:prompt}) instructs the model to, first, determine the reasoning path needed for answering, then to use the \emph{Fact} command for getting relevant facts from the KB, and finally to answer based on the proofs acquired with the \emph{Fact} command. Additionally, we instruct the model to determine which kind of answer is required, to respond ``I don't know.'' when no useful facts have been found, to understand when it is not able to find useful facts and stop, to strictly adhere to the prompt, and to be aware of the description predicate (often useful with the Wikidata KB).
At the end of this prompt, we add two examples to better guide the model behavior. While two examples are not enough to show the model all the possible kind of questions, increasing the number of few-shot example may impact the efficiency of our approach. Therefore, while the approach can work with more demonstrations, we stick to a two-shot prompt in this paper.

During inference, we detect when the LLM generates the sequence of tokens corresponding to the \emph{Fact} command and we activate constrained generation; at this point, the model is forced to generate an existing fact. After an entire fact is generated, we switch the model back to normal generation, allowing it to continue reasoning or to call again the \emph{Fact} command. An example of ReFactX is visible in Figure~\ref{fig:example}.


\section{Experimental Setup}
\label{sec:experiments}

\textbf{Underlying Models.}
\label{sec:exp_qa}
We evaluate ReFactX on the QA task with the following models:
\textit{meta-llama/Llama-3.3-70B-Instruct}~\cite{qwen2025qwen25technicalreport}, \textit{microsoft/phi-4}~\cite{abdin2024phi4technicalreport}, \\ \textit{Qwen/Qwen2.5-72B-Instruct}, and \textit{Qwen/Qwen2.5-7B-Instruct}~\cite{qwen2025qwen25technicalreport}
with the 800M fact tree derived from Wikidata indexed in PostgreSQL (see Section~\ref{sec:wikidata-repr}).





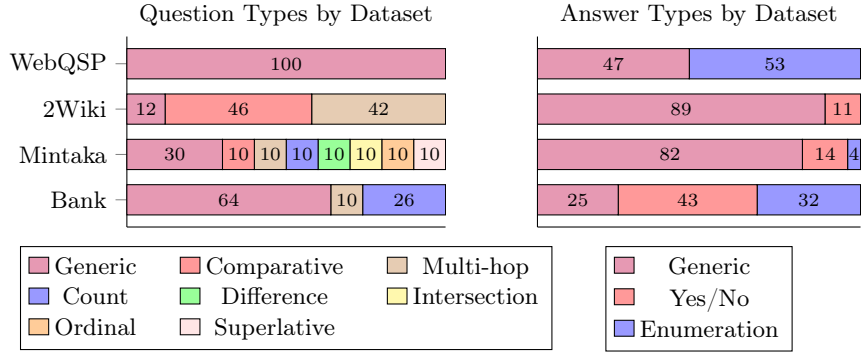
\begin{figure}[ht]
\centering

\begin{minipage}[t]{0.48\textwidth}
\raggedright
\pgfplotsset{testbar/.style={
        title=Question Types by Dataset,
        xbar stacked,
        width=0.99\textwidth,
        height=8cm,
        axis lines*=left,
        xmin=0,xmax=100,
        ytick = data,
        yticklabels = {Bank, Mintaka, 2Wiki, WebQSP},
        xtick=\empty,          
        xticklabels=\empty,   
        tick align = outside,
        bar width=4mm,
        y=6mm,
        enlarge y limits={abs=0.6},
        nodes near coords,
        nodes near coords align={center},
        every node near coord/.style={font=\scriptsize},
        legend style={
            at={(0.5,-0.1)},
            anchor=north,
            legend columns=3,
            /tikz/every even column/.append style={column sep=0.5cm}
        },
        every axis/.append style={
            title style={yshift=-5pt}
        }
    }}
\begin{tikzpicture}[baseline=(current bounding box.north)]
\begin{axis}[testbar]
    \addplot[fill=purple!40] coordinates {(64,0) (30,1) (12,2) (100,3)};

    
    \addplot[fill=red!40] coordinates {(0,0) (10,1) (46,2) (0,3)};
    
    \addplot[fill=brown!40] coordinates {(10,0) (10,1) (42,2) (0,3)};
    
    \addplot[fill=blue!40] coordinates {(26,0) (10,1) (0,2) (0,3)};
    
    \addplot[fill=green!40] coordinates {(0,0) (10,1) (0,2) (0,3)};
    
    \addplot[fill=yellow!40] coordinates {(0,0) (10,1) (0,2) (0,3)};

    \addplot[fill=orange!40] coordinates {(0,0) (10,1) (0,2) (0,3)};
    
    \addplot[fill=pink!40] coordinates {(0,0) (10,1) (0,2) (0,3)};
    
    \legend{
        Generic,
        Comparative,
        Multi-hop,
        Count,
        Difference,
        Intersection,
        Ordinal,
        Superlative,
    }
\end{axis}
\end{tikzpicture}
\end{minipage}
\hfill 
\begin{minipage}[t]{0.48\textwidth}
\centering
\pgfplotsset{testbar/.style={
        title=Answer Types by Dataset,
        xbar stacked,
        width=\textwidth,
        height=8cm,
        axis lines*=left,
        xmin=0,xmax=100,
        ytick = \empty,
        yticklabels = \empty,
        xtick=\empty,          
        xticklabels=\empty,   
        tick align = outside,
        bar width=4mm,
        y=6mm,
        enlarge y limits={abs=0.6},
        nodes near coords,
        nodes near coords align={center},
        every node near coord/.style={font=\scriptsize},
        legend style={
            at={(0.5,-0.1)},
            anchor=north,
            legend columns=1,
            /tikz/every even column/.append style={column sep=0.5cm}
        },
        every axis/.append style={
            title style={yshift=-5pt}
        }
    }}
\begin{tikzpicture}[baseline=(current bounding box.north)]
\begin{axis}[testbar]
    \addplot[fill=purple!40] coordinates {(25,0) (82,1) (89,2) (47,3)};

    \addplot[fill=red!40] coordinates {(43,0) (14,1) (11,2) (0,3)};
    
    \addplot[fill=blue!40] coordinates {(32,0) (4,1) (0,2) (53,3)};
    
    \legend{
        Generic,
        Yes/No,
        Enumeration,
    }
\end{axis}
\end{tikzpicture}
\end{minipage}

\caption{\textbf{Question and answer type distribution across the four evaluation datasets.}
Stacked bars show how each benchmark (Bank, Mintaka, 2Wiki, WebQSP) varies in its mix of question categories---generic, comparative, multi-hop, and others---and answer forms (generic, yes/no, enumeration).}\label{fig:dataset_distribution}
\end{figure}

\subsubsection{Datasets}
We evaluate our approach on three public benchmark datasets---Mintaka~\cite{sen-etal-2022-mintaka}, 2WikiMultiHopQA~\cite{ho-etal-2020-constructing}, WebQSP~\cite{yih-etal-2016-value}---and on an anonymized proprietary financial dataset, referred to as \textit{Bank}.
Mintaka~\cite{sen-etal-2022-mintaka} is a multilingual dataset containing nine question types annotated by crowdworkers with Wikidata IDs; in this work, we only consider eight question types, and only English questions. This dataset allows us to analyze ReFactX's performance across diverse question types.
We treat Yes/No---which is considered a question type in Mintaka---as an answer type, alongside \textit{Generic} and \textit{Enumeration}.
Figure~\ref{fig:dataset_distribution} shows the distribution of questions and answer types across all datasets. 

2WikiMultiHopQA~\cite{ho-etal-2020-constructing}, to which we refer as 2WikiMH, is composed of multi-hop, comparative, and generic questions derived from Wikipedia and Wikidata. For this dataset, we evaluate on the \textit{dev} set, as the ground truth for the test set is not publicly available.
WebQSP~\cite{yih-etal-2016-value} contains generic questions annotated with Freebase~\cite{freebase}. 
These two dataset are used by the existing KE-QA approaches based on constrained generation~\cite{li2024decodinggraphsfaithfulsound,luo2024graphconstrainedreasoningfaithfulreasoning}, providing insights into ReFactX’s performance relative to prior work.
For computational efficiency, we consider a limited sample of 200 questions for each of these datasets, stratifying on the question type.

The proprietary dataset---called \textit{Bank}, as it is from a large European central bank---covers the financial domain, enabling us to study ReFactX's adaptation to domain-specific data.
It has 278 template questions built from an anonymized corporate KG ($\sim10K$ triples, 9 relations such as ownership and control). 
In this setting, we can evaluate LLMs while minimizing data contamination issues, as the models are unlikely to have been exposed to this data during training. 
We verbalize the triples, also adding inverted facts, so that ReFactX can use tail-to-head reasoning. Finally we build an in-memory prefix-tree index. Since in this dataset each relation is considered independent from the other ones, we add specific instructions in the prompt to consider each relation independently.

\subsubsection{Metrics}
\label{sec:metrics}
We evaluate on \textit{Accuracy} (A) and \textit{Precision} (P):

\begin{subequations}
\begin{minipage}{0.48\textwidth}
\begin{equation}
    A = \frac{|\text{Correct Answers}|}{|\text{Questions}|},
\end{equation}
\end{minipage}%
\begin{minipage}{0.48\textwidth}
\begin{equation}
    P = \frac{|\text{Correct Answers}|}{|\text{Given Answers}|},
\end{equation}
\end{minipage}
\end{subequations}

the accuracy (A) corresponds to the ratio of questions answered correctly. The precision (P), instead, is normalized on the number of given answers, excluding ``I don't know'' and answers not given when the allowed maximum number of new tokens have been reached. Intuitively, P represent how precise ReFactX is when it answers.

Precision and accuracy are calculated in two settings: 1) Exact Match: comparing the predicted answer with the ground truth with case-insensitive string equality; 2) LLM-as-a-Judge~\cite{zheng2023llmjudge}: we ask Llama3.3-70B in 16-bit precision whether the predicted answer is correct and complete with respect to the ground truth.

\subsubsection{Reference Approaches}
We first evaluate the same LLMs used with ReFactX without constrained generation. In this setting, referred as \textit{LLM-only}, we use the same settings and prompts, so that models generates not-grounded facts based solely on their parametric-knowledge, giving us clues on how important is to access external knowledge and, furthermore, on how much each dataset can be answered using only the LLM’s internal knowledge.

Then, for each dataset, we consider one reference approach from literature: Hybrid-QA (HQA)~\cite{lehmann2024beyond}---an input-based KE-QA approach---for the Mintaka dataset~\cite{sen-etal-2022-mintaka},
Decoding on Graphs (DoG)~\cite{li2024decodinggraphsfaithfulsound} for 2WikiMultiHopQA~\cite{ho-etal-2020-constructing}, and Graph-Constrained Reasoning (GCR)~\cite{luo2024graphconstrainedreasoningfaithfulreasoning} for WebQSP~\cite{yih-etal-2016-value}. For these approaches, we report the results from the respective papers. HQA evaluation is performed on a 200-question sample of Mintaka, DoG and GCR evaluate on bigger portion of the datasets: for 2WikiMH DoG uses 6964 questions; for WebQSP respectively DoG and GCR considers 1542 and 1628 questions (filtered during pre-processing).

HQA~\cite{lehmann2024beyond} achieves state-of-the-art results on Mintaka~\cite{sen-etal-2022-mintaka}. 
Given a question, HQA first selects a limited set of few-shot examples maximizing their relevance for the question and the diversity between the examples, then, via ICL with the selected few-shot examples, instructs the LLM to acquire external information with tools, such as a Wikipedia search engine and a Wikidata SPARQL query engine, and finally answers the question.

DoG~\cite{li2024decodinggraphsfaithfulsound} and GCR~\cite{luo2024graphconstrainedreasoningfaithfulreasoning} are both based on constrained generation, while they consider smaller question-based prefix trees with respect to ReFactX.
DoG~\cite{li2024decodinggraphsfaithfulsound},
incorporates constrained generated triples from a KG inside a reasoning process similar to CoT, instructing the model with ICL, and alternating normal and constrained generation.
The KG, composed of up to 120 triples, is constructed for each question, from the triples within 2-4 hops from question entities. Then, at inference time, a query-centric subgraph of the KG is obtained from the triples containing the question entities (extracted with an Entity Linking system). The model is allowed to generate only the triples from this query-centric subgraph, and at each generation, the subgraph is expanded with all the adjacent triples, allowing the model to form a reasoning path from question entities to the answer.

GCR~\cite{luo2024graphconstrainedreasoningfaithfulreasoning}, instead, considers a pre-built subgraph of Freebase~\cite{freebase} of 8 million triples containing the entities mentioned in the evaluation questions, then for each questions it constructs a smaller prefix tree from the paths obtained with breadth first search within 2-hops from question entities (obtained with Entity Linking). At this point a LLM, fine-tuned for the task, generates multiple reasoning paths from the prefix tree with constrained decoding and an answer hypothesis for each path. Finally, a powerful LLM, such as GPT-4o-mini, receives all the paths and the hypotheses answers and gives the final answer.

%
DoG and HQA use accuracy in the evaluation, while GCR calculates \textit{Hit} and \textit{F1}. Considering \textit{enumeration} answer (containing a list of items), \textit{Hit} counts a prediction as correct whenever any item of the ground truth matches the prediction, whereas \textit{F1} is the average of the \textit{F1} of the single predictions. These measures are equivalent to accuracy for generic answers (not for enumerations).




%
In our experiments, we use beam search~\cite{hu-etal-2015-improved} with $\text{number of beams} = 3$ (GCR uses 10 beams, ReFactX and DoG 3), we disable sampling, and we set 1000 as the limit of new tokens to generate.


\section{Results and Discussion}
\label{sec:results}

\begin{figure}[h]
    \includegraphics[width=\linewidth]{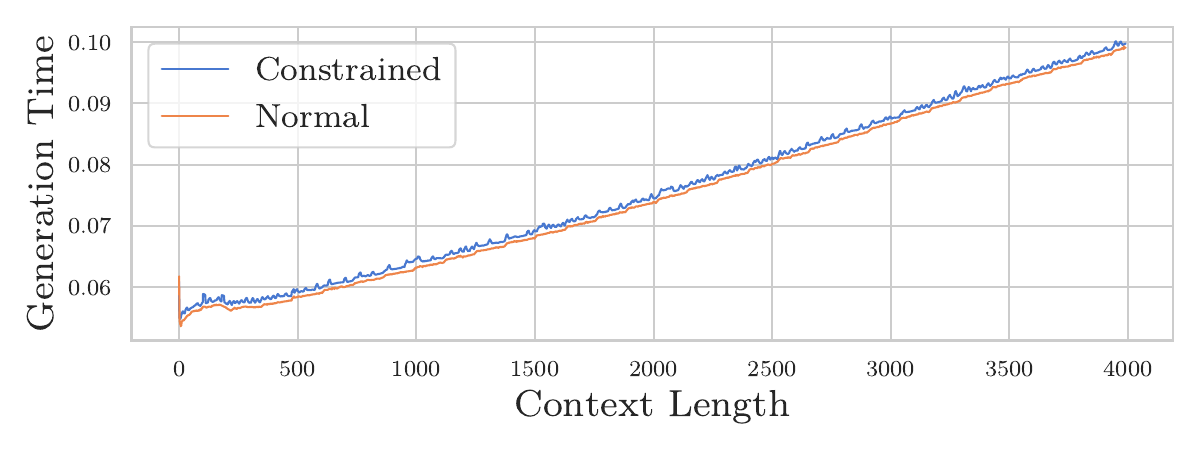}
    \caption{\textbf{Measured token generation time (in seconds).} Results on 4000 tokens, using Qwen2.5-3B with PostgreSQL running on the same machine, with key-value caching enabled, using one Nvidia Tesla T4. We apply a moving average with a 10-tokens window to reduce noise.}
    \label{fig:generation_time}
\end{figure}

\textbf{Generation-Time Overhead.}
\label{sec:exp_overhead}
We compare ReFactX's KB-guided constrained generation time with unconstrained \textit{LLM-only} generation time.
%
Figure~\ref{fig:generation_time} plots the token generation times calculated in the two settings. The time overhead added by constrained generation is very limited: the total time for generating 4000 tokens increases by only 1.3\%, from 300.14 seconds to 303.89.
\begin{table}[ht]
\caption{Experimental results for detecting data contamination, comparing ReFactX with LLM-only baselines across four dataset and considering four different LLMs (details in Section~\ref{sec:exp_qa}). We report Precision (P) and Accuracy (A), calculated with Exact Match (left) and LLM-as-a-Judge (right). }
\label{tab:em_judge_results_memorization}
    \resizebox{\textwidth}{!}{%
\begin{tabular}{lll|ll|ll|ll|ll|ll|ll|ll}
\toprule
 & \multicolumn{8}{c|}{\textbf{Exact Match}} & \multicolumn{8}{c}{\textbf{LLM-as-a-Judge}} \\
 & \multicolumn{2}{c|}{Bank} & \multicolumn{2}{c|}{Mintaka} & \multicolumn{2}{c|}{2WikiMH} & \multicolumn{2}{c|}{WebQSP} & \multicolumn{2}{c|}{Bank} & \multicolumn{2}{c|}{Mintaka} & \multicolumn{2}{c|}{2WikiMH} & \multicolumn{2}{c|}{WebQSP} \\
 & P & A & P & A & P & A & P & A & P & A & P & A & P & A & P & A \\
\midrule
\multicolumn{16}{c}{\textbf{ReFactX}} \\
$\text{Llama}^{3.3}_{70B}$ & \textbf{40.8} & 36.0 & \textbf{66.4} & 40.5 & \textbf{74.4} & 64.0 & \textbf{22.8} & 17.0 & 50.0 & \textbf{42.8} & \textbf{91.8} & 56.0 & 93.6 & 81.0 & 85.2 & 63.5 \\
$\text{Qwen}^{2.5}_{72B}$ & 39.9 & \textbf{37.1} & 43.8 & 28.0 & 71.9 & \textbf{69.0} & 18.5 & 14.0 & 46.1 & \textbf{42.8} & 82.8 & 55.5 & \textbf{96.4} & \textbf{92.5} & 84.8 & 65.5 \\
$\text{Phi}^{4}_{14B}$ & 35.8 & 29.9 & 33.0 & 19.0 & 62.2 & 46.0 & 15.4 & 10.5 & \textbf{51.3} & 41.7 & 88.7 & 51.0 & 92.6 & 69.5 & \textbf{89.7} & 61.0 \\
$\text{Qwen}^{2.5}_{7B}$ & 24.7 & 20.9 & 48.5 & 24.5 & 58.5 & 46.5 & 17.0 & 12.0 & 44.8 & 35.3 & 78.2 & 39.5 & 91.2 & 73.0 & 78.7 & 55.5 \\
\midrule
\multicolumn{16}{c}{\textbf{LLM-only}} \\
$\text{Llama}^{3.3}_{70B}$ & 23.1 & 18.3 & 60.7 & \textbf{59.5} & 46.7 & 43.0 & 19.9 & \textbf{19.5} & 23.1 & 18.3 & 83.7 & \textbf{82.0 }& 61.4 & 56.5 & 82.1 & \textbf{80.5} \\
$\text{Qwen}^{2.5}_{72B}$ & 30.0 & 29.9 & 52.3 & 51.5 & 35.7 & 35.5 & 13.6 & 13.5 & 30.0 & 29.9 & 81.2 & 80.0 & 45.2 & 45.0 & 73.7 & 73.0 \\
$\text{Phi}^{4}_{14B}$ & 21.8 & 20.1 & 43.6 & 42.5 & 25.5 & 24.0 & 14.4 & 14.0 & 25.7 & 23.7 & 76.9 & 75.0 & 41.5 & 39.0 & 75.3 & 73.0 \\
$\text{Qwen}^{2.5}_{7B}$ & 21.7 & 18.0 & 45.3 & 41.0 & 21.6 & 18.0 & 10.9 & 10.5 & 28.1 & 21.9 & 65.7 & 60.0 & 35.9 & 30.0 & 66.7 & 64.0 \\

\bottomrule
\end{tabular}}
\end{table}

\begin{table}[h!]
    \caption{Insights into comparison with Related Work on Precision (P) and Accuracy (A). Results for related work are taken from their papers. ReFactX is evaluated using LLM-as-a-Judge$^\dagger$, HQA via manual annotation$^\ddagger$. DoG and GCR use exact match$^*$. GCR results are calculated with different metrics$^{\S}$ (see Section~\ref{sec:metrics}).}
    \label{tab:res-comparison-rw}
    \centering
\begin{tabular}{l|cc|cc|cc}
\toprule
 & \multicolumn{2}{c|}{\textbf{Mintaka}} & \multicolumn{2}{c|}{\textbf{2WikiMH}} & \multicolumn{2}{c}{\textbf{WebQSP}} \\
 & P & A & P & A & P & A \\
\midrule
\textbf{ReFactX} $\text{Llama}^{3.3}_{70B}$$^\dagger$ & 91.8 & 56.0 & 93.6 & 81.0 & 85.2 & 63.5 \\
\textbf{ReFactX} $\text{Qwen}^{2.5}_{72B}$$^\dagger$ & 82.8 & 55.5 & 96.4 & \textbf{92.5} & 84.8 & 65.5 \\
\textbf{ReFactX} $\text{Qwen}^{2.5}_{7B}$$^\dagger$ & 78.2 & 39.5 & 91.2 & 73.0 & 78.7 & 55.5 \\
\midrule
DoG\cite{li2024decodinggraphsfaithfulsound}$^*$ $\text{Qwen}^{2.5}_{7B}$ & -- & -- & -- & 84.2 & -- & \textbf{92.7} \\
GCR\cite{luo2024graphconstrainedreasoningfaithfulreasoning}$^*$ $\text{Llama}^{3.1}_{8B} + \text{GPT}^{\text{4o-mini}}$ & -- & -- & -- & -- & 92.2$^{\S}$ & 74.1$^{\S}$ \\
HQA\cite{lehmann2024beyond} GPT$^{3.5}$$^\ddagger$ & -- & 85.9 & -- & -- & -- & -- \\
HQA\cite{lehmann2024beyond} GPT$^4$$^\ddagger$ & -- & \textbf{95.9} & -- & -- & -- & -- \\
\bottomrule
\end{tabular}
\end{table}



%

%

\textbf{Performance Analysis.}
\label{sec:performance}
In Table~\ref{tab:em_judge_results_memorization} we show ReFactX results on the benchmark datasets, compared with \textit{LLM-only} results. The evident difference between Exact Match and LLM-as-a-Judge results likely derives from comparing prediction and ground truth with string equality: answers with correct dates in different formats, or enumerations with the correct items in different order are counted as errors with the exact match.
%

Comparing ReFactX with \textit{LLM-only}, we observe that ReFactX always outperforms \textit{LLM-only} models in \textit{precision} in the LLM-as-a-Judge evaluation.
In terms of \textit{accuracy}, ReFactX performs worse than \textit{LLM-only} on the Mintaka and WebQSP datasets. This reveals these datasets are covered by the models parametric knowledge up to a large extent -- reaching 82.0\% and 80.5\% accuracy respectively.
%
%
Instead with 2WikiMH, 
whose questions seem harder for the LLM parametric knowledge,
ReFactX improves both \textit{accuracy} and \textit{precision} by more than 20\% with respect to \textit{LLM-only} models.

The Bank dataset proved especially challenging, with ReFactX achieving precision up to 51.3\% and accuracy up to 42.8\%. However, a type-wise analysis reveals good results on generic Yes/No questions achieving precision (P) of 85.2\% and accuracy (A) of 78.9 (using LLM-as-a-Judge) with Qwen2.5-72B while with Llama3.3-70B we achieves P=77.8\% and C=70.6\%. Our proposed approach is, instead, particularly suffering with count questions: both models achieve less than 10\% accuracy on these questions. Enumeration questions, similarly, prove challenging with only 30\% being answered correctly on the Bank dataset.


With respect to the Mintaka dataset, ReFactX struggles especially with superlative and count questions. With Llama3.3-70B, accuracy reaches just 14.3\% for superlatives and 55.0\% for counts. Qwen2.5-72B shows a contrasting pattern, achieving higher accuracy on superlatives (47.6\%) but lower on counts (35.0\%).
On comparative and multi-hop questions, instead, Qwen2.5-72B achieves 52.4\% (comparative) and 57.9\% (multi-hop) accuracy, while Llama3.3-70B reaches 66.7\% and 73.7\% -- still below the performance levels in 2WikiMH, composed mostly by comparative and multi-hop questions. 

On WebQSP, enumeration answers prove particularly difficult in terms of precision for ReFactX with both Llama3.3-70B and Qwen2.5-72B, while when using Phi-4 it maintains consistent precision across all answer types.

Results from Table~\ref{tab:res-comparison-rw}, showing ReFactX with reference approaches, suggest that our approach can achieve competitive results. In fact, compared to DoG on 2WikiMH---although the two approaches are not evaluated on the same dataset sample---ReFactX, even if considering a large KB, achieves 73.0\% accuracy with Qwen2.5-7B and 92.5\% with Qwen2.5-70B, while DoG stands in the middle with 84.2\%.


\subsubsection{Discussion}



The generation-time overhead measurements along with the results on the benchmark QA datasets demonstrate that LLMs with constrained generation and a prefix tree can effectively access external knowledge without any additional model, retriever, or external service. Additionally, by storing the prefix tree on disk via a database service we are able to scale to large knowledge base of 800M facts. These advantages come with the negligible cost of a $\sim1\%$ increment in generation time.

In particular, ReFactX shows greater effectiveness on the 2WikiMH dataset. This can be explained by the nature of the dataset that contains only generic, comparative, and multi-hop questions, easier to address with our approach (see Figure~\ref{fig:dataset_distribution}), and requires only simple answers. Indeed, answering these question types generally requires fewer facts with respect to count or enumeration questions.
%
While for Mintaka multi-hop questions, we notice that in some cases they require ordinal reasoning, such as \textit{``Where was the 16th president of the United States born?''}, making them harder for ReFactX.

On Mintaka and WebQSP, the datasets widely covered by the internal knowledge of the tested LLMs, when comparing ReFactX behavior with \textit{LLM-only} on the same questions, ReFactX still demonstrates interesting reasoning patterns. In some cases, it uses facts to prove an answer coming from LLM parametric knowledge.
In others, e.g., for \textit{``Where did Rick Santorum attend high school?''}, in which \textit{LLM-only} fails, ReFactX is able to acquire correct facts which lead to correcting model parametric knowledge.


However, ReFactX has some intrinsic limitations. The main one derives from the autoregressive left-to-right nature of LLMs. Indeed, ReFactX requires left-to-right facts, that start with known-information and lead to desired information.
Furthermore, while we believe ReFactX is particularly useful to access point-wise factual information, count questions like \textit{``How many movies have been directed by Danny Boyle?''} or 
\textit{``Which NHL team has the most Stanley Cup wins?''} 
are particularly challenging, because they require ReFactX to enumerate a large list of facts and to understand when all the required facts have been generated.

This limitation likely explains why our approach is obtaining worse performance on count and superlative question types. Such question types, could be better handled by empowering LLMs with additional tools like SPARQL engines, which support counts or other set operations, or by investigating mechanisms that control ReFactX generation similarly as we did for preventing fact repetition.

The obtained results are particularly promising considering that we did not fine-tune the models to improve their ability to leverage ReFactX during reasoning. We plan to address this limitation in future work.

\section{Conclusion and Future Work}
\label{sec:conclusion}
In this work, we presented ReFactX, a generic KB-constrained decoding wrapper that can slot into any autoregressive LLM and any tokenized knowledge base compiled into a prefix tree. By allowing only continuations that lead to valid Knowledge Base facts, it injects point-wise evidence while adding just 1.3\% latency for generating 4K tokens and lifts QA accuracy by up to 20pp at over 90\% precision, compared to relying solely on the LLM's internal knowledge. ReFactX is thus a lightweight, easy-to-integrate route for fact grounding.

We plan to explore fine-tuning for improving LLMs reasoning capabilities with ReFactX and to
extend it to cover counts, long enumerations, and other set operations.
To reach top-tier accuracy there, one can let the model first explore with ReFactX and then pass the harvested entities/predicates to stronger tools e.g., auto-generated SPARQL queries against an endpoint which we leave it for the future work.


\section{Supplemental Material Statement}
Source code for using ReFactX and reproducing our work is available from
GitHub at https://github.com/rpo19/ReFactX.




\section*{Acknowledgments}
This work was supported by the EU Horizon Europe programme (grants No. 101189771---DataPACT and No. 101070284---enRichMyData), the European Community – Next Generation EU via the Italian Ministry of Justice, and the Italian PRIN project Discount Quality for Responsible Data Science (202248FWFS).
The authors also acknowledge the computing resources provided by the NHR Center at TU Dresden, funded by the German Federal Ministry of Education and Research and participating state governments (www.nhr-verein.de/unsere-partner).

%
%
%
\bibliographystyle{splncs04}
\bibliography{mybibfile}
\end{document}